\documentclass[10pt,twocolumn,letterpaper]{article}

\usepackage{iccv}
\usepackage{times}
\usepackage{epsfig}
\usepackage{graphicx}
\usepackage{amsmath}
\usepackage{amssymb}
\usepackage{adjustbox}
\usepackage{float}
\usepackage{amsfonts}
\usepackage{booktabs}
\usepackage{algorithm}
\usepackage{subcaption}
\usepackage{mathtools}
\usepackage{url}            
\usepackage{nicefrac}       
\usepackage{microtype}      
\usepackage{xcolor}         
\usepackage{algpseudocode}

\newcommand{\patmat}{PATMAT\xspace}
\newcommand{\inp }{inpainting\xspace}
\iccvfinalcopy
\usepackage{ wasysym }
\usepackage[capitalize]{cleveref}
\crefname{section}{Sec.}{Secs.}
\Crefname{section}{Section}{Sections}
\Crefname{table}{Table}{Tables}
\crefname{table}{Tab.}{Tabs.}

\begin{document}

\title{PATMAT: Person Aware Tuning of Mask-Aware Transformer for Face \inp}

\author{Saman Motamed \\
CMU\\{\tt\small sam.motamed@insait.ai}
\and
Jianjin Xu\\
CMU
\and Chen Henry Wu\\
CMU
\and Fernando de la Torre \\
CMU
}
\twocolumn[{%
\renewcommand\twocolumn[1][]{#1}%
\maketitle
\begin{center}
    \centering
    \captionsetup{type=figure}
    \includegraphics[width=0.90\linewidth]{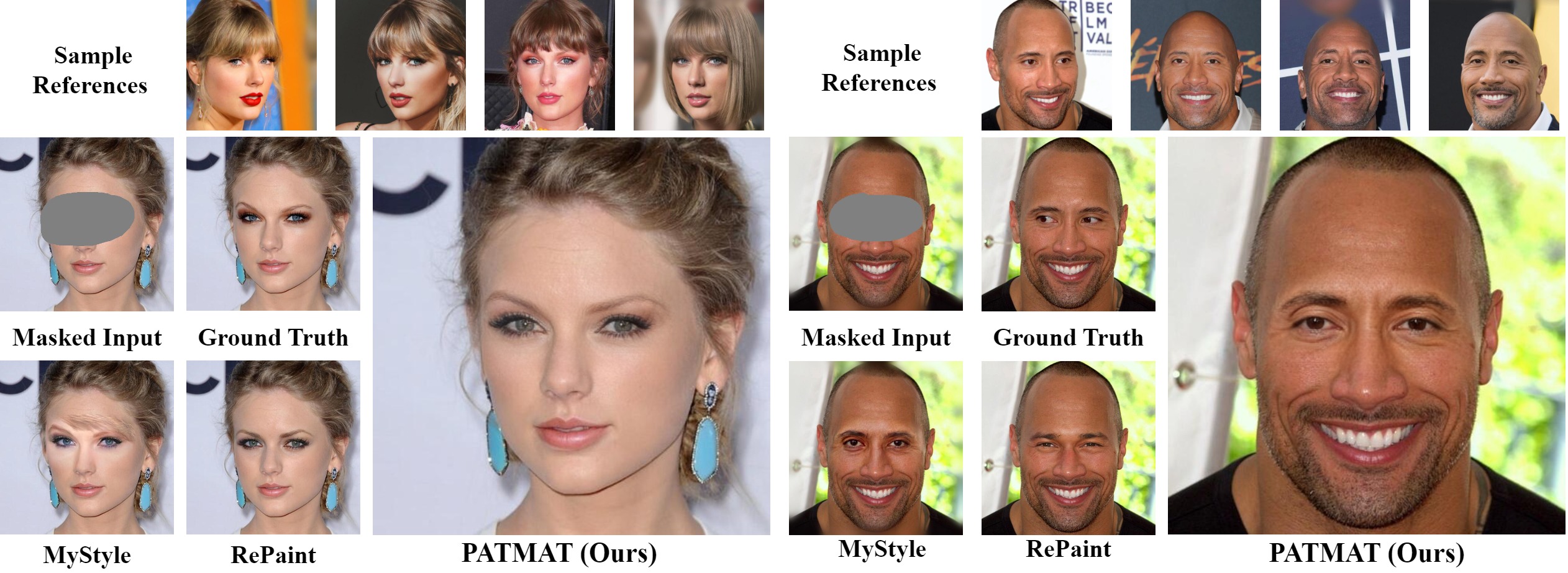}
    \captionof{figure}{\label{fig:first} An illustration of \patmat's identity preserving inpainting. By using reference images of Taylor Swift and Dwayne Johnson, PATMAT enables Mask-Aware Transformer to preserve the identity of the person. We also show results of recent methods such as  MyStyle~\cite{nitzan2022mystyle}, trained with the same number of reference images as \patmat and RePaint\cite{lugmayr2022repaint} that is not fine-tuned with any reference images.}
\end{center}%
}]
\maketitle
\ificcvfinal\thispagestyle{empty}\fi
\begin{abstract}
Generative models such as StyleGAN2 and Stable Diffusion have achieved state-of-the-art performance in computer vision tasks such as image synthesis, inpainting, and de-noising. However, current generative models for face inpainting often fail to preserve fine facial details and the identity of the person, despite creating aesthetically convincing image structures and textures. In this work, we propose Person Aware Tuning (PAT) of Mask-Aware Transformer (MAT) for face inpainting, which addresses this issue.
Our proposed method, PATMAT, effectively preserves identity by incorporating reference images of a subject and fine-tuning a MAT architecture trained on faces. By using \(\sim 40\) reference images, PATMAT creates anchor points in MAT's style module, and tunes the model using the fixed anchors to adapt the model to a new face identity. Moreover, PATMAT's use of multiple images per anchor during training allows the model to use fewer reference images than competing methods.
We demonstrate that PATMAT outperforms state-of-the-art models in terms of image quality, the preservation of person-specific details, and the identity of the subject. Our results suggest that PATMAT can be a promising approach for improving the quality of personalized face inpainting.

\end{abstract}

\section{Introduction}
\label{sec:intro}

The objective of image inpainting is to generate plausible content to complete missing regions within an image.
Preserving contextual integrity of the inpainted image is a crucial factor, where the reconstructed regions must conform to reasonable structure and texture based on local and non-local priors in the image. This task becomes increasingly challenging when addressing larger and irregular missing regions in the image. In particular, facial inpainting poses a significant challenge as it requires maintaining fine facial details and the subject's identity, which are essential for various applications including security (e.g., inpainting a face behind a mask or sunglasses), entertainment (e.g., seeing a person while they are wearing a virtual reality headset), or photo restoration. In each of these tasks, it is essential to keep the subject's identity and fine facial details intact. However, recent inpainting methods have largely focused on generating high-quality images with little attention given to preserving the identity of the subject. Therefore, we explore the feasibility of inpainting techniques that aim to maintain the subject's identity and facial characteristics while generating high-quality, photo-realistic images.

Recent works have pushed the boundaries of large-hole image \inp \cite{MAT, wan2021high, suvorov2022resolution, ma2022regionwise}.
Mask-Aware Transformer (MAT) used vision transformers, with a multi-head contextual attention mechanism with shifting windows \cite{liu2021swin} to build long-range dependency priors and achieved great \inp results compared to competing methods such as CoModGAN \cite{zhao2021large}, ICT \cite{wan2021high} and LaMa \cite{suvorov2022resolution}. While these models are able to synthesize high quality inpainted images, the reconstruction result sometimes fails to recover the details of the ground truth.
Particularly, in the case of face inpainting, existing models (e.g., MAT, CoModGAN, RePaint) cannot recover the same subtle facial details as the ground truth (e.g., brows shape, eye look, beard shape,...). As a result, they do not preserve the subject's identity after inpainting (see Fig.~\ref{fig:first}).
For tasks such as image editing and restoration, especially in the domain of faces, identity preservation is a requirement, yet the problem of identity preserving image \inp is relatively unexplored.
\par This work proposes \patmat for personalized face \inp. Given a few reference images of a person, PATMAT integrates this information into a pre-trained MAT model by tuning the network parameters, conditioned on the style vectors within MAT. We compare two style conditioning methods and propose a regularization loss to prevent over-fitting to the reference images during the tuning process.
To evaluate our method, we curated images of seven public figures due to the lack of diverse and sufficiently large datasets (no public datasets have sufficient number of images per identity such as CelebA-HQ \cite{karras2017progressive} and VFHQ \cite{xie2022vfhq}). Our qualitative and quantitative experiments demonstrate that PATMAT outperforms several contemporary state-of-the-art inpainting models in terms of quality and identity-preservation.
\newline\noindent \textbf{Contributions:}
\begin{enumerate}
\item We propose PATMAT, a tuning method based on creating anchors in MAT's style space that allows for high quality personalized \inp of the face.
\item We propose a regularization method that controls over-fitting during tuning and further improves the quality of personalized \inp.
\item In order to make our personalized method more practical, our effort reduces the number of required reference images from \(\sim 100-200\) in previous works \cite{nitzan2022mystyle} to \(\sim 40\) images.
\item  We used images of seven celebrities for our experiments to overcome the limitations of existing datasets.

\end{enumerate}
\section{Related Work}
\label{sec:relwork}
Image \inp has been a long standing task in computer vision. Early pixel-matching and diffusion-based \inp methods \cite{barnes2009patchmatch, bertalmio2000image, bertalmio2003simultaneous}, which propagated neighbouring pixel information to the masked region lacked a high level understanding of the image that hindered them from generating semantically reasonable images and have since been replaced by more sophisticated deep learning approaches.

\par Pathak \etal's \cite{pathak2016context} use of an encode-decoder architecture, equipped with adversarial training and a pixel-wise reconstruction loss objective achieved photo-realistic \inp results and has since been the basis of more follow-up works \cite{hui2020image, liu2020rethinking, ntavelis2020aim, zeng2019learning, ren2019structureflow}. 

With recent advances of diffusion models in image synthesis \cite{kingma2021variational, dhariwal2021diffusion, ruiz2022dreambooth, nichol2021glide, saharia2022palette}, RePaint \cite{lugmayr2022repaint} leveraged the expressiveness of a pre-trained Denoising Diffusion Probabilistic model \cite{ho2020denoising} and used it as a prior in their \inp model.

Success of GANs in synthesizing high quality images \cite{karras2019style, Karras_2020_CVPR} brought forward different methods and variations \cite{liu2021pd, demir2018patch}, adapting them for image inpainting. Most GAN-based \inp methods are prone to deterministic transformations due to limited control during synthesis \cite{wu2022generative}. UTCGAN \cite{zhao2020uctgan} and Zhao \etal \cite{PIC2019} proposed VAE-based networks to combat this issue. By projecting images and masked counterparts onto a low-dimensional manifold space and optimizing the KL-divergence, UTCGAN \cite{zhao2020uctgan} achieved pluralistic generation capabilities for mask filling. CoModGAN \cite{zhao2021large} used a co-modulation layer in order to improve reconstruction and diversity of their inpainting.

\par To the best of out knowledge, PTI \cite{roich2022pivotal} and MyStyle\cite{nitzan2022mystyle} are the only works that focused on personalizing models for editing out-of-domain (OOD) images. MyStyle built on the work presented by pivot tuning (PTI) of StyleGAN's \cite{karras2019style} latent space. PTI was proposed as a method for fine-tuning a pre-trained Generative Adversarial Network that enabled the model to generate OOD images. Given an OOD image of interest ($x_i$), PTI tunes the Generator to generate $x_i$. First, by projecting \(x_{i}\) onto the pre-trained GAN's latent space, a pivot latent code \(w_i\) is acquired. While \(G(w_i)\) has some characteristics of \(x_{i}\), due to the image being OOD, it cannot fully recover the same identity of the person. PTI, while keeping \(w_i\) unchanged, trains \(G\) to minimize \( \mathcal{L}(G(w_i), x_{i})\) where $\mathcal{L}$ is the reconstruction objective. MyStyle used pivot tuning and $\sim 100 - 200$ reference images in order to create a personalized manifold in the latent space of a StyleGAN and showed identity preserving image editing capabilities such as \inp and super resolution in this manifold. 

PTI and MyStyle work with latent space of GANs and require a large number of reference images to learn a personalized manifold. MAT however, does not operate with a GAN-like latent space. Instead, it uses a noise-style space and a style manipulation module that enables pluralistic generation. We draw inspiration from PTI's use of anchors and condition the style manipulation module by defining anchors in the noise-style space. We show that our method is able to preserve the identity for \inp and achieves high quality image synthesis, competing with recent SOTA \inp models. \patmat's use of multiple images per anchor during training allows the personalization of the model by using as few as $\sim 40$ reference images (see PATMAT-C's improvement in identity preserving, that uses multiple image sper anchor over PATMAT-S that uses one image per anchor in table \ref{tab:quant}).

\section{Method}
\label{sec:method}

Given an image $x$ of person $p$, a binary mask image $b$, and a set of reference images $\mathcal{X}=\{x_i\}_{i=1}^N$ of person $p$, \patmat aims to personalize face \inp by fine-tuning a pre-trained MAT, such that \(x \ \astrosun \ b \) can be filled with plausible content that preserves the identity of $p$.

\subsection{MAT Architecture}

The recently proposed MAT~\cite{MAT} achieved great quality inpaintings of images with large holes, compared to an ensemble of recent models~\cite{zhao2021large, wan2021high, yi2020contextual, zhao2020uctgan}. MAT used vision transformer blocks~\cite{vit} to model long range dependencies in the image. In addition, inspired by \cite{Karras_2020_CVPR}, MAT introduced a style manipulation module that enabled pluralistic mask filling for a single masked image. 

Given an input image $x$, a binary mask image $b$, and an unconditional noise-style code $\textbf{s}_u \in \mathbb{R}^d$, MAT generates an inpainted image $\hat{x} = \text{MAT} (x \ \astrosun \ b, \textbf{s}_u)$.
$\textbf{s}_u$ is used in formulating \textbf{s} (\ref{eq:s}), that is tasked with manipulating MAT's weight normalization of convolution layers in the image reconstruction layers, such that using different $\textbf{s}_u$ inputs results in different inpaintings of the masked image input.
To enhance the representation ability of $\textbf{s}_u$, MAT fuses the noise-style code $\textbf{s}_u$ with an image-conditional style code $\textbf{s}_c$. $\textbf{s}$ and $\textbf{s}_c$ are formulated as:

\begin{gather} \label{eq:styles}
    \textbf{X}^{'} = \textbf{B} \ \astrosun \ \textbf{X} + (1 - \textbf{B}) \ \astrosun \ \text{Resize}(\textbf{s}_u) \\
    \label{eq:style_c}
    \textbf{s}_c =  \mathcal{F}(\textbf{X}^{'})
    \\
    \label{eq:s}
    \textbf{s} = \mathcal{A}(\textbf{s}_u, \textbf{s}_c)
\end{gather}
Where $\mathcal{F}$ and $\mathcal{A}$ are mapping functions, \textbf{B} is a random binary mask and $\textbf{s}_c$ is conditioned on both the image features $\textbf{X}$ (output of transformer blocks given image $x$) and the unconditional noise-style $\textbf{s}_u$. 
The resulting style code $\textbf{s}$ then changes the inpainted image by manipulating the weights $\textbf{W}$ of convolution layers in the reconstruction network:
\begin{equation}\label{eqn:weighted_conv}
\textbf{W}'_{ijk} = \textbf{W}_{ijk} \ . \ \textbf{s}_i
\end{equation}
where $i, j, k$ denote the input channels, output channels and spatial footprint of the convolution. \patmat uses the unconditional noise-styles $\textbf{s}_u$ in order to personalize MAT.
\label{method: matoverview}

\subsection{Person Aware Tuning}
PATMAT allows tuning of MAT through its style manipulation framework using a few reference images, enabling the \inp of faces while preserving the identity. We draw inspiration from PTI's \cite{roich2022pivotal} idea of fixing a pivot in a GAN's latent space and tuning the generator to generate an out-of-distribution (OOD) image using the fixed pivot.

\patmat conditions the $\textbf{s}_u$ noise-style codes by treating them as anchors and fixing them during training, such that the inpainted image $\hat{x}_i = MAT_{\theta} (x_i \ \astrosun \ b, \mathcal{A}(\textbf{s}_{c_i}, \textbf{s}_{u_i}))$ minimizes the reconstruction objective formulated as:
\begin{equation}\label{eqn:loss}
\mathcal{L}_{\theta}= \mathcal{L}_P + \mathcal{L}_2 (x_i, \hat{x}_i)  
 \end{equation}
 where ${\theta}$ represents the parameters of MAT, $\mathcal{L}_P = ||\phi_i (\hat{x_i}) - \phi_i(x_i)||_1$ is the perceptual loss \cite{fid}, $\phi_i(.)$ denotes the layer activation of a pre-trained VGG-19~\cite{simonyan2014very} network and $\mathcal{L}_2$ is the Mean Squared Error.

\begin{figure}
\centering
    \includegraphics[width=0.75\linewidth]{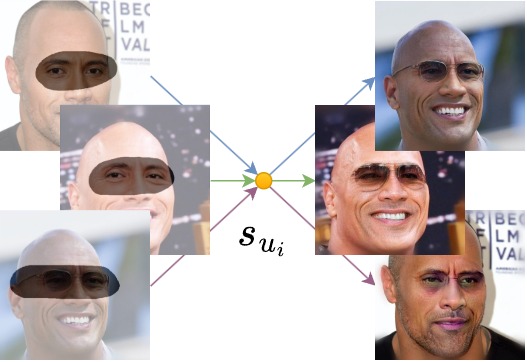}
\caption{\label{fig:leak} An illustration of how using the same pivot to inpaint images with and without sunglasses leads to the features of sunglasses leaking to other images.}
\end{figure}

\subsubsection{Style Conditioning}
\par We propose two different methods for tuning MAT, using reference images $x_i \in \mathcal{X}$ and corresponding noise-style anchors $\textbf{s}_{u_i}$. \patmat-S uses a single image per anchor in its tuning mechanism (similar to how PTI uses one latent code per image) while \patmat-C uses multiple images per anchor. We explored two methods (random vs optimized) for calculating initial anchors in section \ref{find-style} and found both to be equally effective.

{\bf PATMAT-S.} For each reference image $x_i$ and its corresponding anchor \(\textbf{s}_{u_i} \in \mathcal{S}_u\), while keeping the anchor fixed, we tune MAT such that inpainted image $\hat{x}_i$ minimizes $\mathcal{L} (x_i, \hat{x}_i({\theta}))$ (\ref{eqn:loss}),
\begin{equation}
    \theta = \text{argmin}_\theta \sum_{i=1}^N \mathcal{L}(x_i, \text{MAT}_\theta (x_i \odot b, \mathcal{A}(\mathbf{s}_{u_i}, \mathbf{s}_{c_i})),
\end{equation}
where $\theta$ is the parameters of MAT.

While this improved identity preserving ability of MAT for inpainting, we found that tuning with multiple images per anchor better utilizes the limited number of reference images and further improved the \inp by taking advantage of the ability that MAT's style manipulation module has in using a single $\textbf{s}_u$ to inpaint multiple images.

{\bf PATMAT-C.} Observe that MAT can use multiple reference images $x_j$ per style anchor $\textbf{s}_{u_i}$. In other words, we can tune the network using $\mathcal{L}(x_i, MAT_\theta (x_i \odot b, \mathcal{A}(\mathbf{s}_u, \mathbf{s}_{c_i}))$. The main difference with PATMAT-S is that in PATMAT-C, the $\mathbf{s}_u$ anchors stay the same for all the reference images. Unlike latent vectors in a GAN's latent space that are unique per each image, MAT's style code \textbf{s} (\ref{eq:s}) is comprised of both the noise-style anchor $\textbf{s}_{u_i}$ and an image-conditioned style $\textbf{s}_{c_i}$. Given two images $x_i , x_j \in \mathcal{X}$, $\textbf{s}_{u_i}$ can be used to inpaint both images since $\textbf{s}_{c_i}$  will be different from $\textbf{s}_{c_j}$. Instead of tuning MAT to inpaint each image using its unique style anchor, we can use each style anchor \(\textbf{s}_{u_i} \in \mathcal{S}_u\) and enforce the network to generate all reference images \(x_j \in \mathcal{X}\) using that single style anchor.

\begin{figure}
\centering
    \includegraphics[width=0.7\linewidth]{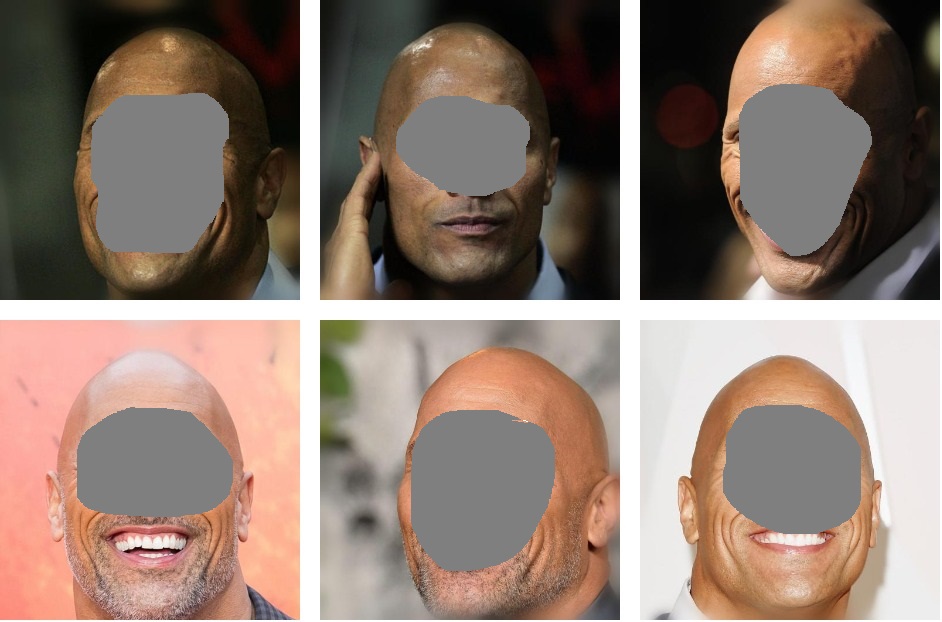}
\caption{\label{fig:lighting} An illustration of how different lighting conditions in the training images can drastically affect the skin-tone of the person.}
\end{figure}


However, enforcing MAT to reconstruct every image \(x_j \in \mathcal{X}\) using a single anchor $s_{u_i}$ is too restrictive. We observed that during tuning, using the same anchor to inpaint images with distinct accessories will lead to features of those accessories propagating to other images. Figure \ref{fig:leak} shows an example of this phenomena where an anchor $\textbf{s}_{u_i}$ is used to inpaint both images with and without sunglasses, leaving the inpainted results with glasses-like shadow over the eyes where there are no sunglasses in the ground truth. This kind of effect can also happen to images with drastic difference in skin tone, as shown in figure \ref{fig:lighting}.
To mitigate this type of effect, we perform a manual grouping of the reference images. Most importantly, we separate images with glasses and sunglasses, each in their own group. Furthermore, in the case of having images with noticeably different skin tone as shown in figure \ref{fig:lighting}, we group such images as well. This step not only improves the \inp results, but also speeds up \patmat by reducing the number of image - anchor pairs used in training.
This results in \(M\) groups of reference images, $\mathcal{X}_{m}$, and the anchor for each group, $\mathcal{S}_{u_m} = \{ \mathbf{s}_{u_m} \}$.
As a whole, we have \(\mathcal{X} = \cup_{m=1}^{M} \mathcal{X}_{m}, \ \text{and} \ \mathcal{S}_u = \cup_{m=1}^{M} \mathcal{S}_{u_m} \).
In our experiments, $ M \sim 3$. Now, for all groups of images \(\mathcal{X}_m\) and their corresponding anchors \(\mathbf{s}_{u_m}\) , we tune MAT by \inp each \(x_i \in \mathcal{X}_m\) using all \(s_{u_i} \in \mathcal{S}_{u_m}\). Our results in table \ref{tab:quant} showed that \patmat-C performed better in preserving the identity compared to \patmat-S and our qualitative study (\ref{res:qual}) further confirmed this finding.

\par Furthermore, by adding random noise to anchors and creating anchor clusters, we observed more stable \inp results over different runs of the algorithm. In this setting, for each group of images \(\mathcal{X}_m\) and their corresponding style anchors \(\mathcal{S}_m\), we fixed \( |\mathcal{X}_m| - 1 \) new anchors around each \(s_{u_i} \in \mathcal{S}_m\) where each new anchor \(s_{u_{i, m}} \ \text{is calculated as} \ s_{u_i} + \epsilon\) with  \(\epsilon \sim \mathcal{N}(0, 1)\). \patmat-C then uses unique anchors for each reference image, such that $x_i$ is inpainted using $s_{u_i}$ and \(s_{u_{j, m = i}} \text{for} \ j \in \{1, 2, ..., |\mathcal{X}_m|\} \ \text{where} \ j \neq i\). PATMAT-C can be described as
\begin{align}
  & \text{argmin}_\theta \sum_{m=1}^{M} \sum_{x_i \in \mathcal{X}_m} \mathcal{L}(x_i, \hat{x}_i(\theta)), \\
  & \hat{x}_i (\theta) = \text{MAT}_\theta (x_i \odot b, \mathcal{A}(\mathbf{s}_{u_{j,m}} + \epsilon_i, \mathbf{s}_{c_i})),
\end{align}
where $\hat{x}_i(\theta)$ is the inpainted image, $\mathbf{s}_{u_{j,m}}$ stands for the $j^{th}$ shared cluster anchor for group $m$, $\epsilon_i \sim \mathcal{N}(\mathbf{0}, \mathbf{I})$ is the noise added for each image, $\mathbf{s}_{c_i}$ is the image conditional style vector for image $x_i$.
\par Note that training without fixed anchors faces the same issues as we showed in figure \ref{fig:leak}. See Appendix \ref{supp:noanchor} for more details.

\subsubsection{Regularization via Anchors}
\label{method: stylereg}

Tuning MAT using fixed anchors exhibits over-fitting similar to those observed in generative models\cite{nitzan2022mystyle, roich2022pivotal}.
This occurs in the form of cascading the features of person $p$ not only local to anchor style codes, but throughout the network.

\begin{figure}
\centering
    \includegraphics[width=0.78\linewidth]{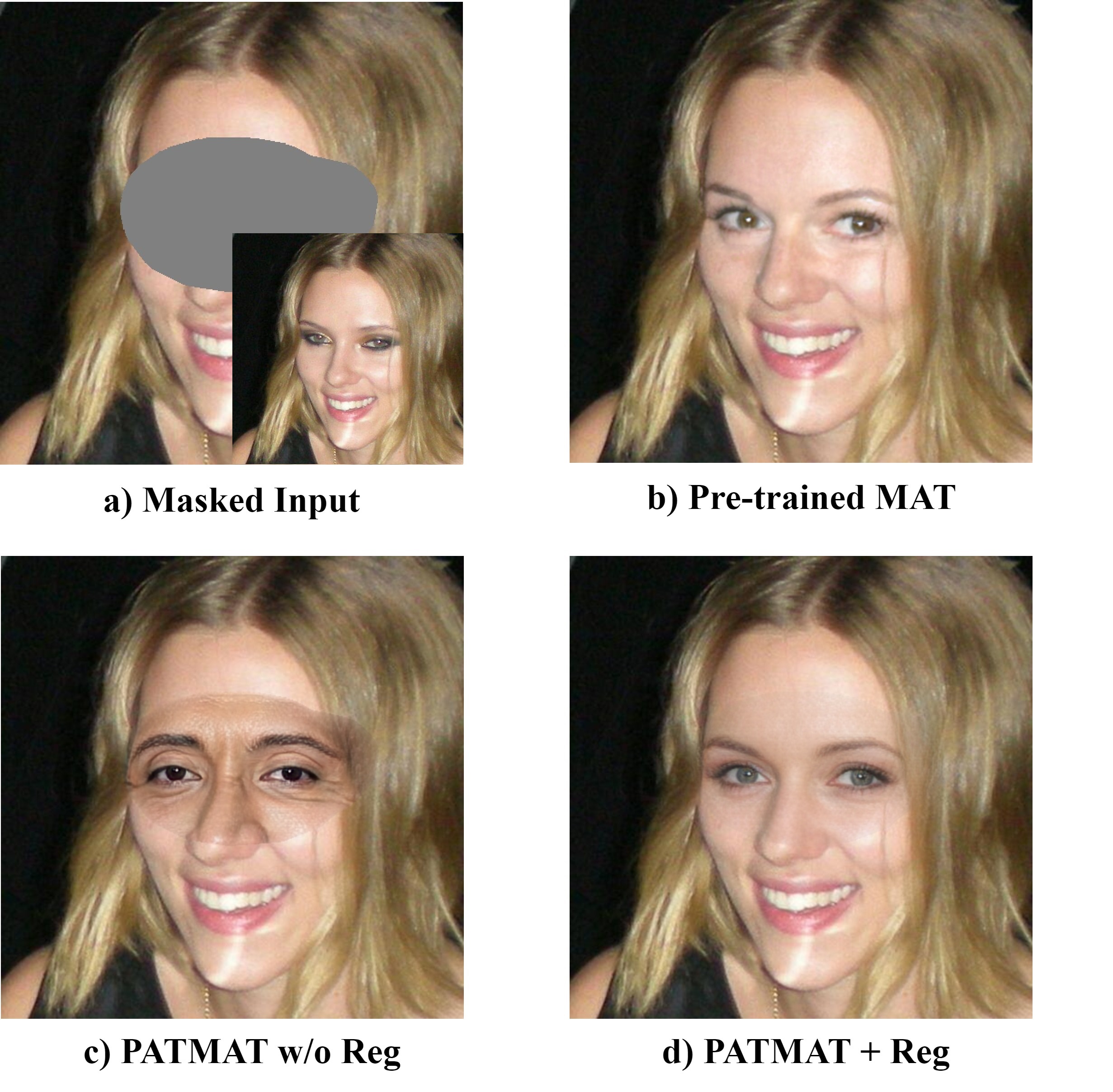}
\caption{\label{fig:ripple}An illustration of PATMAT's over-fitting. \textbf{b} shows a pre-trained MAT's \inp results for a given masked image (\textbf{a}). \textbf{c} and \textbf{d} show the result using the same input \textbf{a}, with \patmat tuned for personalizing images of Dwayne Johnson, without and with regularization, respectively. Note that these images are not a scenario where we would use PATMAT since the image to be inpainted is not of Dwayne Johnson and this figure is only used to illustrate the over-fitting and regularization effects.}
\end{figure}

\begin{figure}
\centering
    \includegraphics[width=1\linewidth]{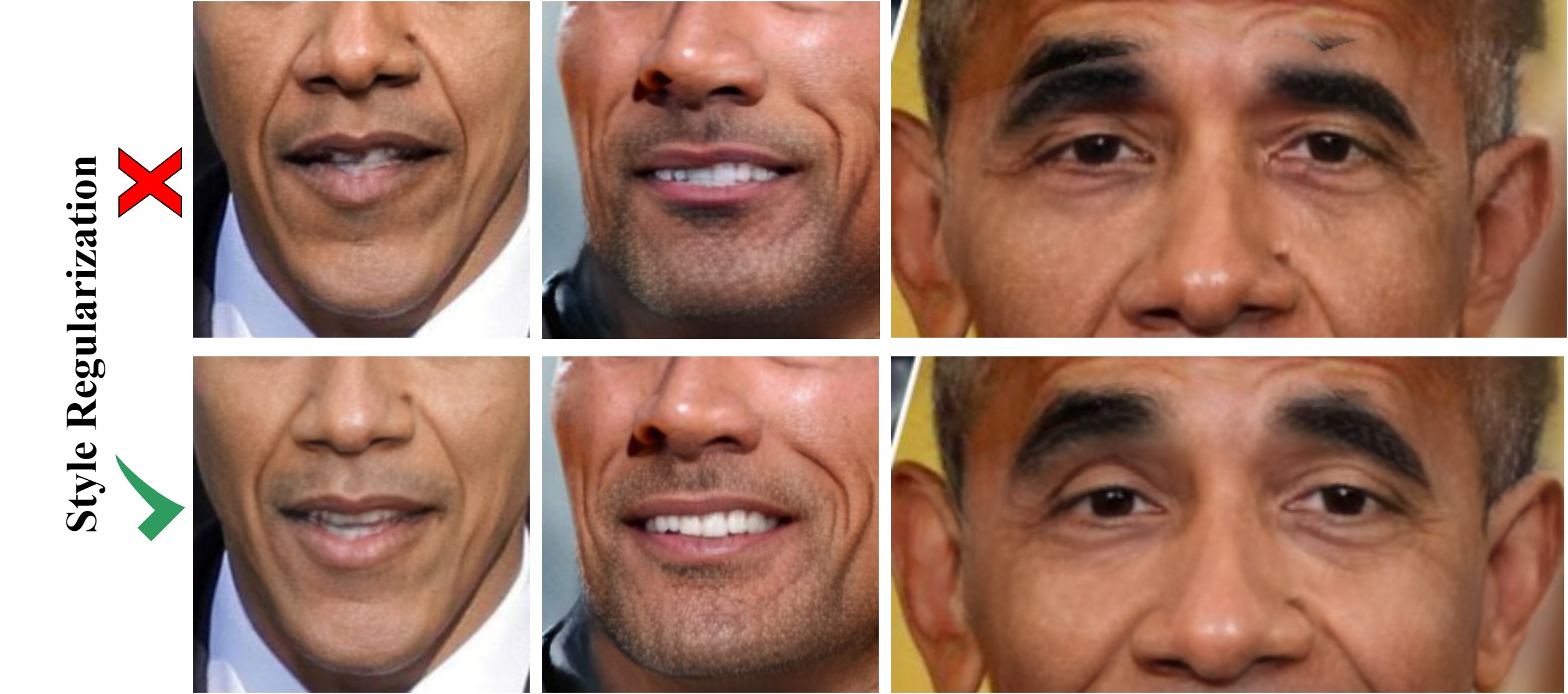}
\caption{\label{fig:reg}  An illustration of the effects of Style Regularization on \patmat's performance. Regularization (bottom row) improves structural integrity of the \inp compared to no regularization (top row).}
\end{figure}

Figure \ref{fig:ripple} - c shows the \inp result of a masked image after \patmat tuning the network with reference images of Dwayne Johnson.
While we want to use \patmat for personalized \inp and do not care about images of other identities being misrepresented, figure \ref{fig:reg} - top row shows examples of this over-fitting interfering with inpainting images of person $p$ where there is misalignment in the inpainted teeth and eyebrows.
To mitigate this, we use a small number ($\sim 30$) of randomly sampled images from MAT's original training data $\mathcal{X}_T$ (or any other set of random images that are not of person $p$). We regularize \patmat using $\mathcal{X}_T$. We fix a unique style anchor $\textbf{s}_{u_t}$ for each ${x}_t \in \mathcal{X}_T$ and at each iteration, minimize $\mathcal{L}_{reg} = \mathcal{L} (\hat{x}_r,  \hat{x}^{'}_{r})$ (\ref{eqn:loss}) where:

\begin{align}
 \hat{x}_r = \text{\patmat}_{\theta^{'}}(x_t \ \astrosun \  b, \ \textbf{s}_{u_t})  \nonumber \\
 \hat{x}_r^{'} = \text{MAT}_{\theta}(x_t \ \astrosun \  b, \ \textbf{s}_{u_t})\nonumber
\end{align}
In other words, we regularize the tuning by enforcing the new PATMAT network to generate the same \inp on a random set of images $\mathcal{X}_T$ that the pre-trained MAT would generate, using the same anchor point. Figure \ref{fig:ripple}-d shows the regularization effect where \patmat tuning of Dwayne Johnson does not interfere with somwhat reasonable \inp of another face. Regularization improves the \inp results for person $p$. Figure \ref{fig:reg} shows the results of two networks, trained for personalizing Barack Obama and Dwayne Johnson. The regularized networks (bottom row) better synthesises color consistency and facial feature alignment compared to the unregularized networks (top row) where there is misalignment in the teeth and eye-brows. 

\par Algorithm \ref{alg:cap} gives an overview of \patmat-C. For \patmat-S, Cartesian products ``$\times$" are replaced by dot products ``$*$".

\begin{algorithm}
\caption{PATMAT}\label{alg:cap}
\begin{algorithmic}
\State initialize $\text{PATMAT}_{\theta^{'}} = \text{MAT}_{\theta}$
\For{$(\mathcal{X}_m, \mathcal{S}_{u_m})$ in $\mathcal{X} \times \mathcal{S}_{u}$}
\For{$(x_i, s_{u_i})$ in $\mathcal{X}_m \times \mathcal{S}_{u_m}$}
\State $\theta^{'} \leftarrow$ $\text{argmin}_{\theta^{'}}$ \( \mathcal{L}  (\text{PATMAT}_{\theta^{'}}(x_i  \astrosun   b,  s_{u_i})\ ,  x_i ) \)
\For{$(x^{'}_j, s^{'}_{u_j})$ in $\mathcal{X}_T * \mathcal{S}_T$}
\State $\theta^{'} \leftarrow$ $\text{argmin}_{\theta^{'}}$ \( \mathcal{L} (\text{PATMAT}_{\theta^{'}}(x^{'}_j  \astrosun b^{'}, s^{'}_{u_j}),\) \par 
 \hskip\algorithmicindent\( \text{MAT}_{\theta}(x^{'}_j  \astrosun  b^{'},  s^{'}_{u_j})) \)
\EndFor
\EndFor
\EndFor
\State return $\text{PATMAT}_{\theta^{'}}$
\end{algorithmic}
\end{algorithm}

\subsection{Inpainting}
\label{method: inpaint}
After tuning the network with \patmat, given a masked image \( x_{ib} = x_{i} \ \astrosun \ b\) to be inpainted, we optimize over noise-style codes in order to refine the inpainting results. While we used a combination of perceptual and $\mathcal{L}_2$ loss to fine-tune the model, pixel-wise metrics are not appropriate for optimizing an identity preserving operations since the same identity can have different facial expressions. We opt to use the ArcFace \cite{deng2019arcface} face identity detection model ($\phi(.)$) instead. We optimize $s_u$ with the objective to minimize the following loss during inpainting:
\begin{gather}
\label{eq:arcface}
    s_u = \text{argmin}_{s_u}  \mathcal{L}_\text{ArcFace}, \\
    \mathcal{L}_\text{ArcFace} = 1 - \cos (\phi (\text{PATMAT}_{\theta^{'}}  (x_{ib}  , 
     s_{u})) , \phi(x_p)),
\end{gather}
where $x_p$ is a random image from reference images $\mathcal{X}$.


\section{Experiments}
\label{experiments}

\begin{figure*}[h]
\centering
    \includegraphics[width=0.82\linewidth]{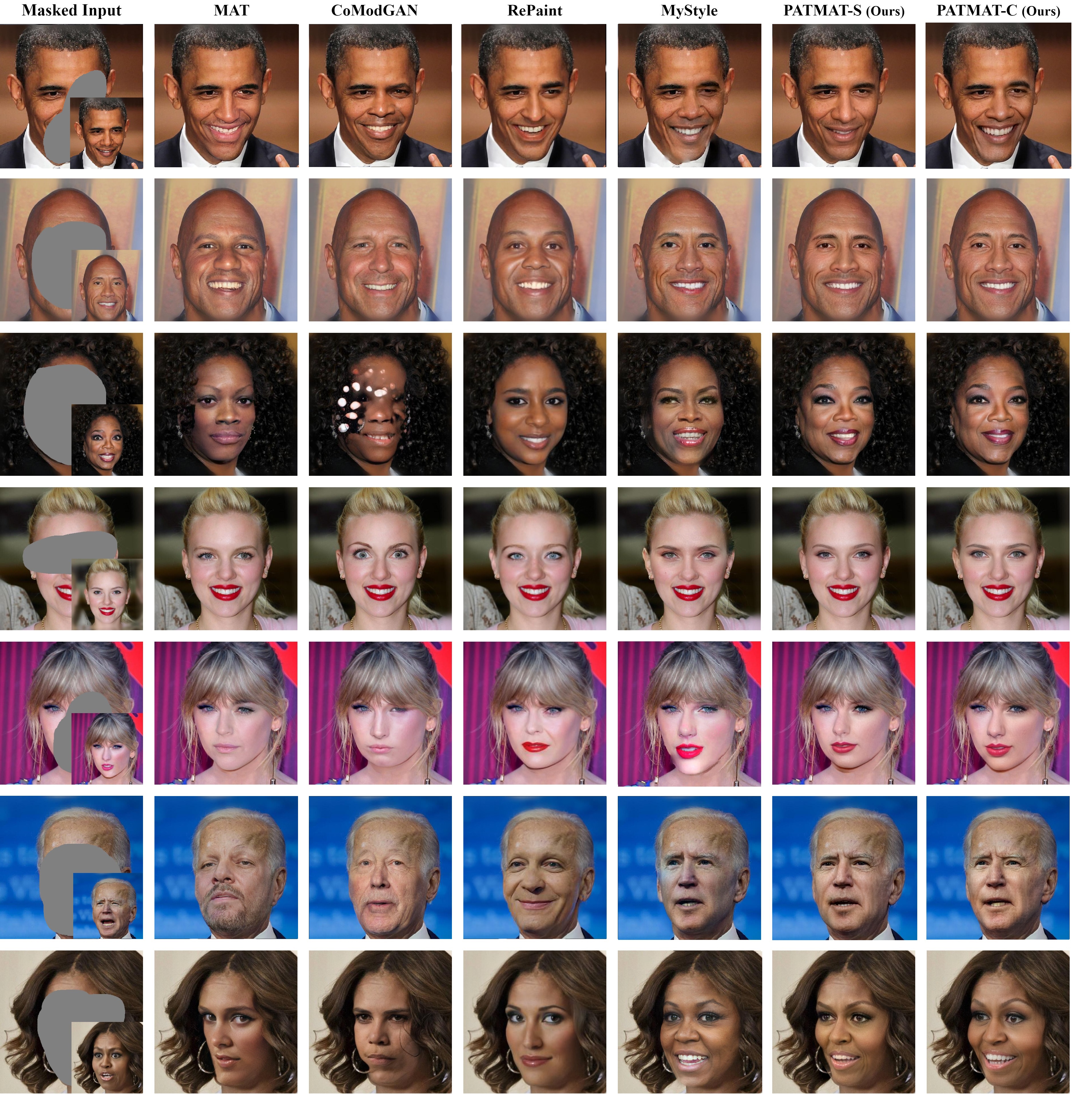}
\caption{\label{fig:qualitative-result} A qualitative comparison of different inpainting models, with images of seven identities used in this study (from top to bottom: Barack Obama, Dwayne Johnson, Oprah, Scarlett Johansson, Taylor Swift, Joe Biden and Michelle Obama). Only \patmat and MyStyle used reference images for tuning. MAT and RePaint were pre-trained on CelebA-HQ while RePaint and MyStyle was pre-trained on FFHQ. Please zoom in to see details.}
\end{figure*}

\subsection{Datasets and Metrics}
There is a lack of high-quality public datasets that could cater to the problem of identity preserving image editing. For this reason, studies resort to using images of celebrities and public figures \cite{nitzan2022mystyle}. However, such images are not representative of day to day photos, since celebrities are often posing and smiling for photos, often in bright light. Due to this limitation, in this paper we show our method's identity preserving \inp performance using images of seven known figures (Michelle and Barack Obama, Joe Biden, Taylor Swift, Oprah, Scarlett Johansson and Dwayne Johnson) (Appendix \ref{supp:dataset}). We selected these well known figures to facilitate our perceptual study, since most people know the looks of these celebrities. 

We evaluated the inpainting results in terms of quality and identity preservation. For quality, we opt to use the \textbf{FID} \cite{fid} score as it has been widely adopted for reporting the quality of generative model samples \cite{Karras_2020_CVPR, MAT, suvorov2022resolution, zhao2021large} and has shown to correlate well with human perception. Pixel-wise \textbf{L1} distance, \textbf{SSIM}\cite{ssim} and \textbf{PSNR} have a weak correlation with human perception regarding image quality \cite{sajjadi2017enhancenet, ledig2017photo}. For identity preservation, we rely on a pre-trained face recognition network, ArcFace \cite{deng2019arcface}. We calculated the cosine similarity of features between the inpainted image and the ground truth. To avoid adversarial attacks on the evaluation model, we used different models in inpainting and evaluation. For evaluation, we used the R50 model trained on the MS1MV3 dataset; for inpainting, we used the R100 model trained on the Glint360K dataset \cite{an2022killing}.

\par Our experiments showed that ArcFace disregards some degree of misalignment and color discrepancy between inpainted areas of the image and the outer mask. For instance, ArcFace would give a high similarity score to \patmat's inpainting of Barack Obama in figure \ref{fig:error}, where a human judge would immediately see the image as abnormal. We further conducted a user study to measure the performance of \patmat with human judges.

 
\subsection{Finding Style Anchors}
\label{find-style}

MAT uses the same mapping network as StyleGAN \cite{Karras_2020_CVPR}, mapping Gaussian noise $\textbf{z}_i$ to noise-style code $\textbf{s}_{u_i}$. \patmat uses an initial style anchor $\textbf{s}_{u_i}$ per reference image $x_i$. We experimented with two methods for obtaining an initial style anchor for each image. 
\subsubsection{Random Style Anchors}
\label{style:rand}
In one set of experiments, for each reference image $x_i$, we picked a random noise $z_i$, mapped to $s_{u_i}$. 
\subsubsection{Optimized Style Anchors}
\label{style:opt}
In the second set of experiments, we optimized $\mathcal{L}_\textbf{z} (x_i, \hat{x}_i)$ such that the noise vector \textbf{z} minimizes the similarity between a reference image and its \inp results, using the pre-trained MAT network and a random mask (similar to projecting images onto a GAN's latent space).

\par We did not observe a significant difference between the \inp performance, using the two different methods for assigning anchors.

\subsection{Implementation Details}
We followed the same architecture design proposed in MAT. Images were resized to $512 \times 512$ resolution and were aligned using a face key-point detection model, consistent with those used in FFHQ \cite{karras2019style} and CelebA \cite{liu2015faceattributes} datasets. The initial calculation of style codes were performed by optimizing $\textbf{z}$ for 200 steps (\ref{style:opt}), using the reconstruction loss described in equation \ref{eqn:loss}. For inpainting, we optimized for $\textbf{s}_{u}$ via ArcFace loss (\ref{eq:arcface}) for 50 steps.
During \inp and quantitative analysis of the \inp results, we used different pre-trained weights for ArcFace. All experiments were carried out on a single NVidia RTX A4000 GPU with 16 GB of memory.

\subsubsection{Poisson Blending}

\begin{figure}
\centering
    \includegraphics[width=0.95\linewidth]{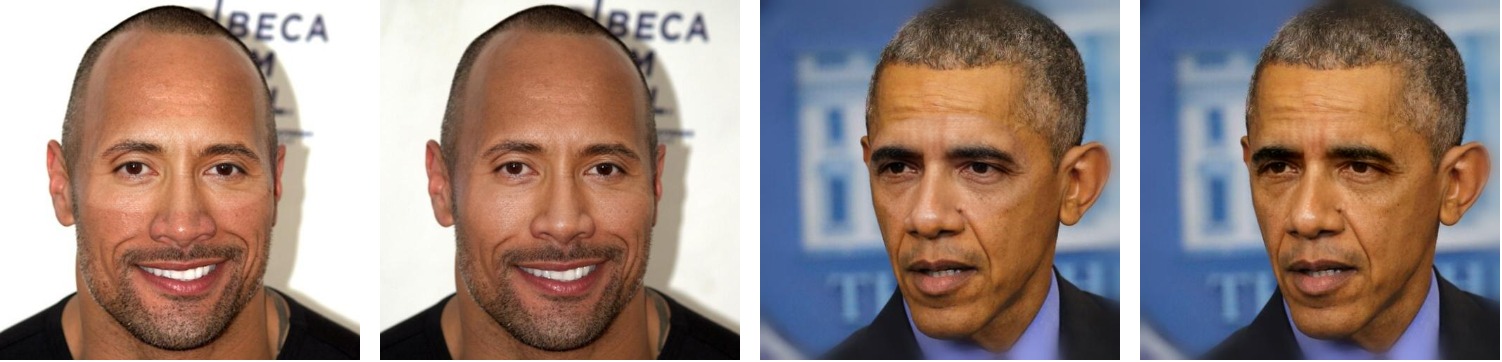}
\caption{\label{fig:poisson} An illustration of Poisson blending where PATMAT's raw output (left) shows RGB inconsistencies in the inpainted image, and blending (right) can mitigate this undesired effect.}
\end{figure}

A caveat of using limited reference images is that the images do not cover the many lighting conditions that affect the skin-tone. For that reason, there could be color inconsistencies between the masked region and the prior. To mitigate this undesired effect, we performed a simple yet effective Poisson blending \cite{perez2003poisson} of the output as shown in figure \ref{fig:poisson}.  

\subsection{Comparison with State of the Art}

We compare the proposed \patmat-C and \patmat-S with a number of contemporary approaches. For a fair comparison, we use publicly available models to test on the same masks. MyStyle \cite{nitzan2022mystyle}, pre-trained on FFHQ \cite{karras2019style} dataset,  is the only model that is tuned besides \patmat, using the same reference images. RePaint \cite{lugmayr2022repaint}, MAT \cite{MAT} and \patmat are trained on CelebA-HQ \cite{liu2015faceattributes} and CoModGAN is trained on FFHQ. 

\subsection{Results}
\subsubsection{Quantitative Analysis}
\label{res:quant}
Table \ref{tab:quant} shows the FID \cite{fid} measurement of Ground Truth (train + test)  images against different \inp results on the test images. MAT achieved the best FID, with \patmat-S and \patmat-C achieving the second and third best FID scores. To measure identity preservation in the inpainted images, we calculated the average cosine similarity of each inpainted image, against all ground truth (train + test) images. \patmat-C achieved the best identity preservation of \textbf{0.67}, with the ground truth images achieving the threshold of 0.70. MAT, CoModGAN and RePaint achieved low ID scores since they are not tuned to preserve identity. While MyStyle does better than these models, with as few as $\sim 40$ reference images as PATMAT uses, it often achieves lower quality inpainting compared to our method (see figure \ref{fig:qualitative-result} and table \ref{tab:quant}).

\subsubsection{Qualitative Analysis}
\label{res:qual}
We conducted a perceptual user study to answer the following two questions:
\begin{itemize}
  \item Which \patmat model (\textbf{S} vs \textbf{C}) better preserves the identity?
      \vspace{-0.2cm}
  \item Are users able to detect images generated by \patmat from real images?
\end{itemize}

\noindent To answer the first question, we randomly selected 5 images per identity (35 total) and used the same mask to inpaint the images with both \patmat-S and \patmat-C. We asked participants to choose the image that most closely resembled the celebrity without providing them with a reference image. The results of the study showed that $62\%$ of the participants preferred PATMAT-C as a better representation of the person compared to PATMAT-S. 

In the second survey, we evaluated the perceptual quality of PATMAT-C's inpainted images by asking human judges to distinguish between real images and inpainted images. The participants were presented with sets of images consisting of three real images and one inpainted image or one real image and one inpainted image per question. The judges were then asked to pick the image they believed was inpainted. 10 questions used three real images and one inpainted image and 10 other questions used one real and one inpainted image per question. Participants were able to pick \patmat's output from three real images with an average $44\%$ accuracy. Given two choices, one real and one inpainted image, the average accuracy increased to $56\%$. In either case, it was close to random chance. 

\noindent We excluded obvious failure cases as shown in figure \ref{fig:error} and randomly picked images from the remaining results. The surveys did not use any repeating images (see Appendix \ref{supp:qual} for more details).

\begin{table}[!ht]
 \small\centering


    \vspace{-0.5cm}
    \begin{tabular}{@{}ccc@{}}

        & \multicolumn{2}{c}{} \\
        \cmidrule(l){1-3}
        & $FID \downarrow$ & $ArcFace \uparrow$ \\
        \cmidrule(l){2-3}
        Ground Truth& -- & 0.70 \\
        MAT \cite{MAT} & $\textbf{22.9}^*$  & 0.35   \\
        MyStyle \cite{nitzan2022mystyle} &24.3  & 0.55   \\
        RePaint \cite{lugmayr2022repaint}&  31.0     &  0.32  \\
        CoModGAN \cite{zhao2021large} &26.0 & c  0.34  \\
        \cmidrule(l){1-3}
        \patmat-S (ours)& $\textbf{23.5}^{\dagger}$ &  $\textbf{0.63}^{\dagger}$   \\
        \patmat-C (ours) & 23.7 & $\textbf{0.66}^*$\\
        \cmidrule(l){1-3}
    \end{tabular} \\
    \caption{Comparison of FID and face identity similarity between the generated images and the ground truth images using \patmat and competing methods.}
            \label{tab:quant}
    \footnotesize{$*$ and $\dagger$ denote the best and second best result respectively}\\
\end{table}

\subsection{Error Analysis}
\begin{figure}
\centering
    \includegraphics[width=0.75\linewidth]{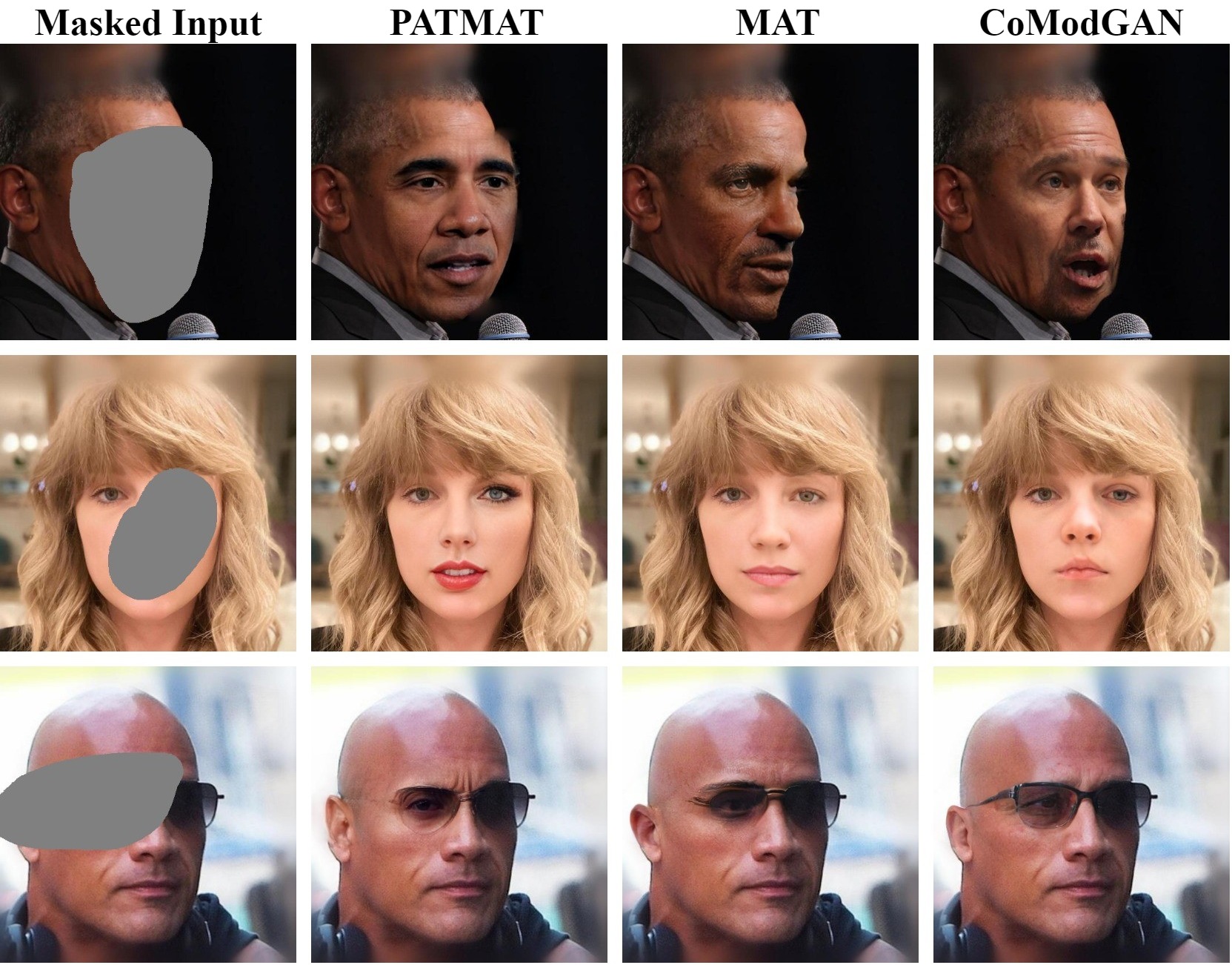}
\caption{\label{fig:error} Examples of \patmat failing to properly inpaint an image. These cases are representative of general types of failures that occur.}
\end{figure}

When the collective of reference images used to train \patmat fails to cover specific poses, accessories, lighting, etc. that appear in an image to be inpainted, \patmat can fail to properly inpaint the image. We showed an example of how lighting affects the results and proposed Poisson blending which works well in mitigating this undesired effect. Other failures however, are not easily fixed without retraining the models with more reference images. Figure \ref{fig:error} shows a few examples of such failures, with CoModGAN and MAT's outputs for comparison. Most of \patmat failed \inp attempts are on images with poses that are close to a side-view of a face (figure \ref{fig:error} - row 1). The second row of figure \ref{fig:error} shows how \patmat inpaints part of Taylor Swift's face with makeup on, while the unmasked part does not have makeup. This is due to almost all the reference images having makeup during tuning. Last row shows how \patmat cannot replicate the shape of the missing sunglasses properly, due to these specific sunglasses not represented in the training data.

\section{Conclusion, Limitation and Future Work}
\label{app:conclusion}
This work proposed PATMAT, a tuning method that takes advantage of Mask-Aware Transformer's multi image per style-code \inp ability. We showed that using as few as $\sim 40$ reference images, we were able to create a personalized MAT that competes with SOTA \inp methods in terms of image quality, while preserving the identity of a person of interest. Our study showed that while PATMAT-S has a slight advantage over PATMAT-C in terms of FID, PATMAT-C outperformed PATMAT-S in qualitative inpainting and identity preservation, as evaluated by the ArcFace model. 

\noindent\textbf{Limitation and future work: \ \ } 
We provided an insight into \patmat's limitations in figure \ref{fig:error}. \patmat reflects the limitations in the reference images it uses to create a personalized \inp space. For instance, we showed that poses that the reference images do not cover, lead to poor \inp results. We are interested in further exploring the role that style codes play in MAT's image synthesis properties. While \patmat showed both random (\ref{style:rand}) and structured (\ref{style:opt}) anchors perform well  preserving the identity, it would be interesting to treat the style-space as a GAN-like latent space and test for different properties such as image editing capabilities. Our method currently relies on a manual data separation step where we separate images with glasses and sunglasses and different lighting conditions. We would like to explore end-to-end methods to automate this step and finally, we would like to generalize \patmat to non-face images (buildings, pets, etc).


\bibliographystyle{IEEEtran}
\bibliography{egbib}

\clearpage
\appendix
\twocolumn[%
   \begin{center}
\textbf{\large Supplemental Materials: PATMAT}

   \end{center}]
   
\section{Dataset}
\label{supp:dataset}
We used images of seven public figures and celebrities to train our models. For evaluation, we used 35 images per person, with the number of training images shown in table \ref{tab1_supp}. Dwayne Johnson and Oprah Winfrey were the only people which had images with glasses and sunglasses. For that reason, we manually split them into train and test sets to have images with glasses in both sets. For the rest of the celebrities, we randomly split them into train and test sets.

\begin{table}[h]
\centering
 \begin{tabular}{||c c c||} 
 \hline
 Celebrity & Training size & Test size  \\ [0.5ex] 
 \hline\hline
 Barack Obama & 47 & 35  \\ 
 Dwayne Johnson & 46 & 35  \\
 Oprah Winfrey & 43 & 35 \\
 Scarlett Johansson & 45 & 35 \\
 Taylor Swift & 41 & 35  \\  
 Joe Biden & 42 & 35 \\
 Michelle Obama & 40 & 35 \\[1ex]
 \hline
 \end{tabular}
 \caption{Training and test sizes used in our experiments.}
 \label{tab1_supp}
\end{table}

\section{Qualitative Study}
\label{supp:qual}

\par We created three surveys as follows. survey \textbf{A} had 19 participants and 35 questions. Each question gave participants the choice to pick an image that best represented one of the seven people we used in our study (table \ref{tab1_supp}). Both given options were inpaintings of the same image, with the same mask, one using PATMAT-S and one using PATMAT-C. We asked 5 questions per identity, totalling 35 questions.
\par These questions relied on the participants knowing and remembering what each figure looked like. We provided 3 real images of each person, not appearing anywhere else in our study and survey questions, as a reference. $62\%$ of the answers picked PATMAT-C as a better representation of the celebrities. 

Survey \textbf{B} had 28 participants and 10 questions. Each question had 4 different images of a person. Each participant had to pick what they thought was an inpainted image from three real images and one inpainted image using PATMAT-C. Survey \textbf{C} followed the same structure as survey \textbf{B}, with 21 participants, except in each question, the participants were given two images of a person; one real and one inpainted. In Survey \textbf{B}, $43\%$ of the answers were able to tell an image was inpainted. In survey \textbf{C}, $56\%$ were able to correctly pick the inpainted image.

\section{Additional Error Analysis}
\label{supp:error}
\begin{figure}
\centering
    \includegraphics[width=0.95\linewidth]{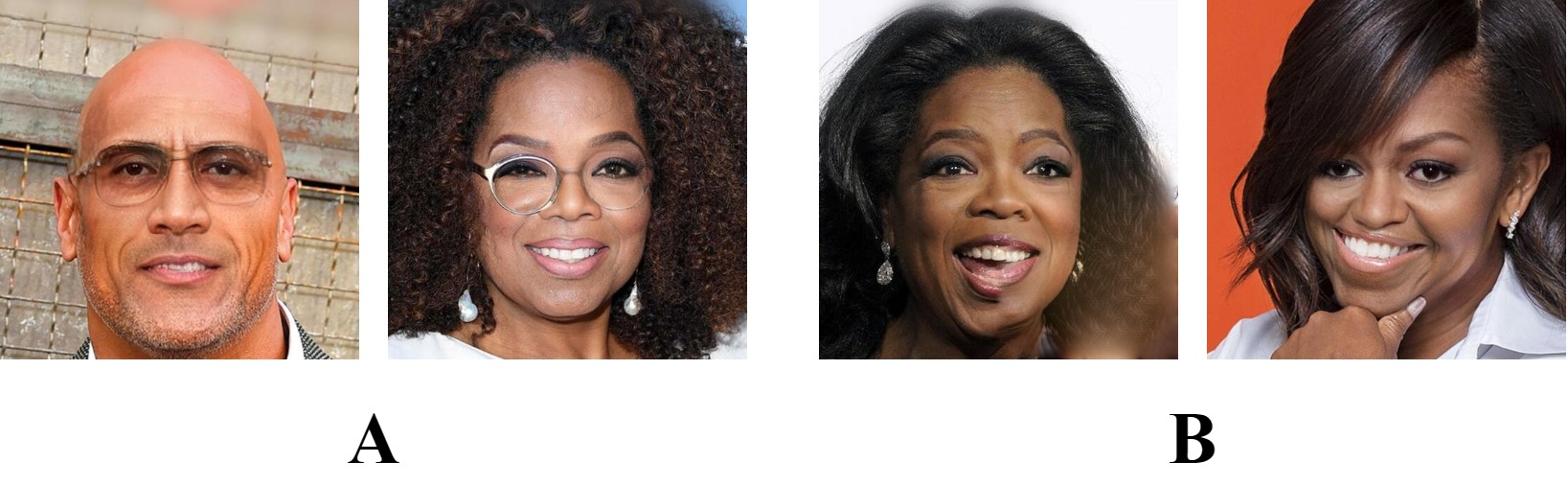}
\caption{\label{figsupp:err} Additional error analysis examples of PATMAT.}
\end{figure}

In figure \ref{figsupp:err}, we show additional scenarios where PATMAT fails to properly inpaint an image. Figure \ref{figsupp:err}-\textbf{A} shows examples where PATMAT attempts to generate / reconstruct sunglasses, yet fails to do so in a convincing fashion. Our experiments showed that this happens when the reference images do not contain the same type of glasses, hence PATMAT inpaints the missing part of the glasses with what is has seen in the references. Figure \ref{figsupp:err}-\textbf{B} shows examples of PATMAT failing to properly align the inpainted area with the local and non-local priors.  

\section{Tuning Without Anchors}
\label{supp:noanchor}
\begin{figure}
\centering
    \includegraphics[width=0.95\linewidth]{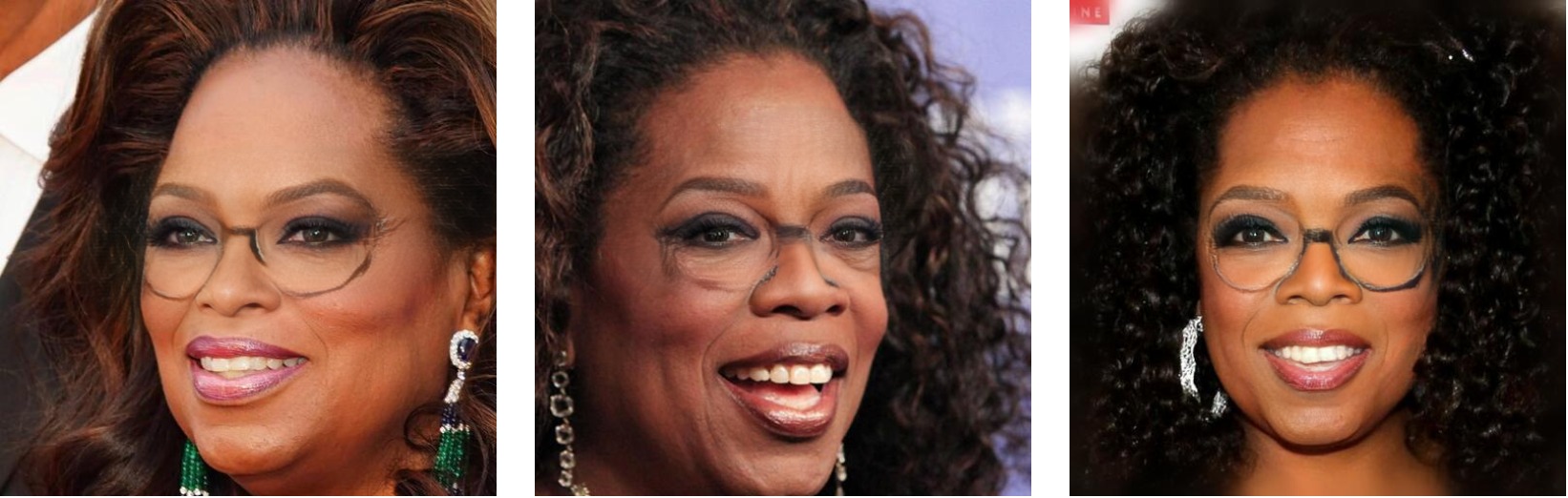}
\caption{\label{figsupp:leak} Tuning MAT without anchors for Oprah, inpaints images with glasses like features.}
\end{figure}

Anchors give structure to tuning MAT. We showed how using one anchor point to inpaint images with and without glasses will result in glasses-like features spreading to other images. Training with no anchors will  have a similar effect, where in each iteration, random $s_u$ noise-style codes are picked for tuning with reference image $x_i$. Without any structure to $s_u$ codes, separating the features, figure \ref{figsupp:leak} shows similar behaviour to what we saw in figure \ref{fig:leak}. 

\section{Additional Qualitative Results}
Figure \ref{more:res} shows additional inpainting examples using PATMAT-S, PATMAT-C and other competing methods.
\onecolumn
\begin{figure}[h]
\centering
    \includegraphics[width=0.99\linewidth]{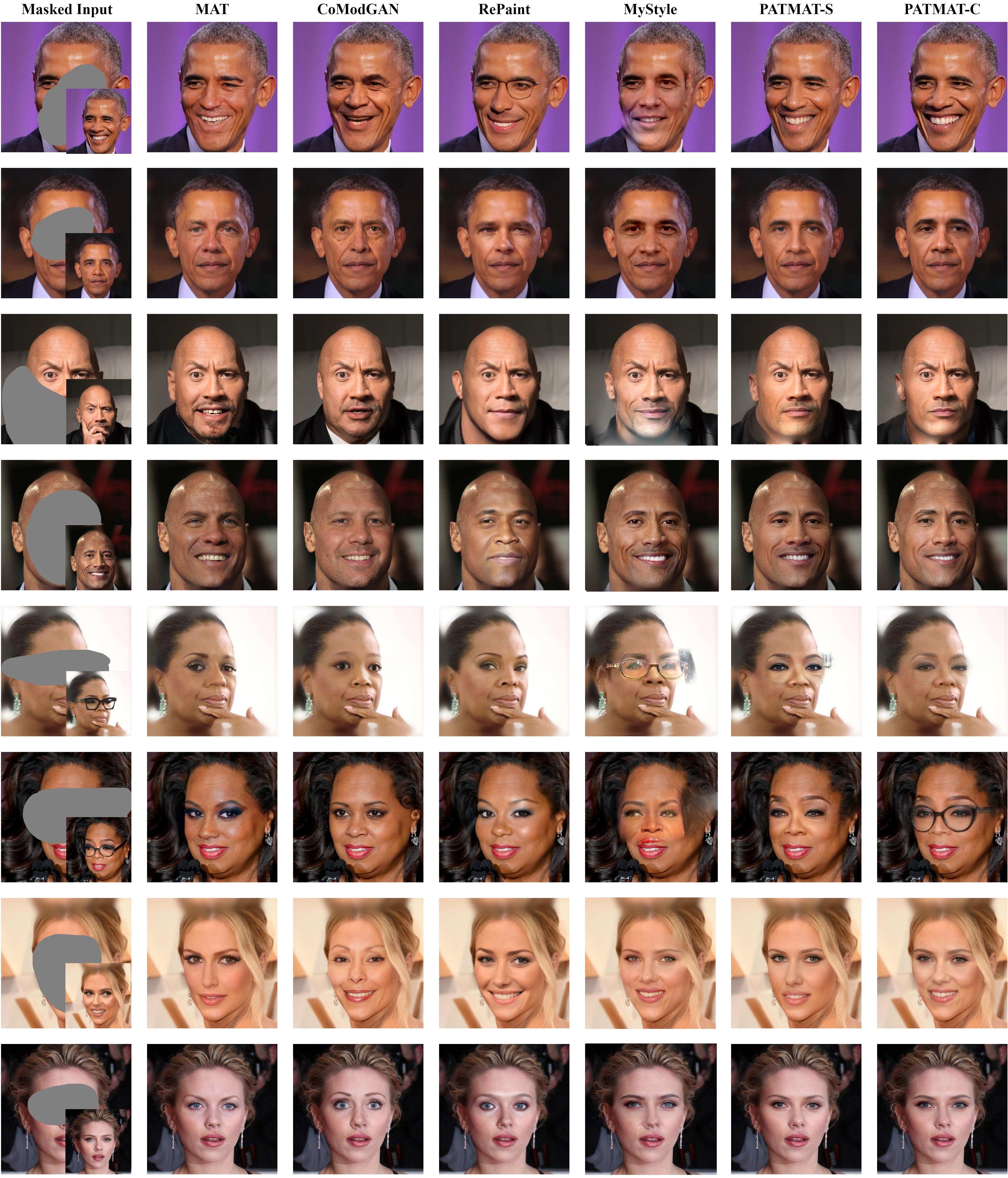}
  \caption{\label{more:res} Additional qualitative results. Please zoom in for details.}
\end{figure}
\begin{figure}[t]
\ContinuedFloat
\centering
    \includegraphics[width=0.99\linewidth]{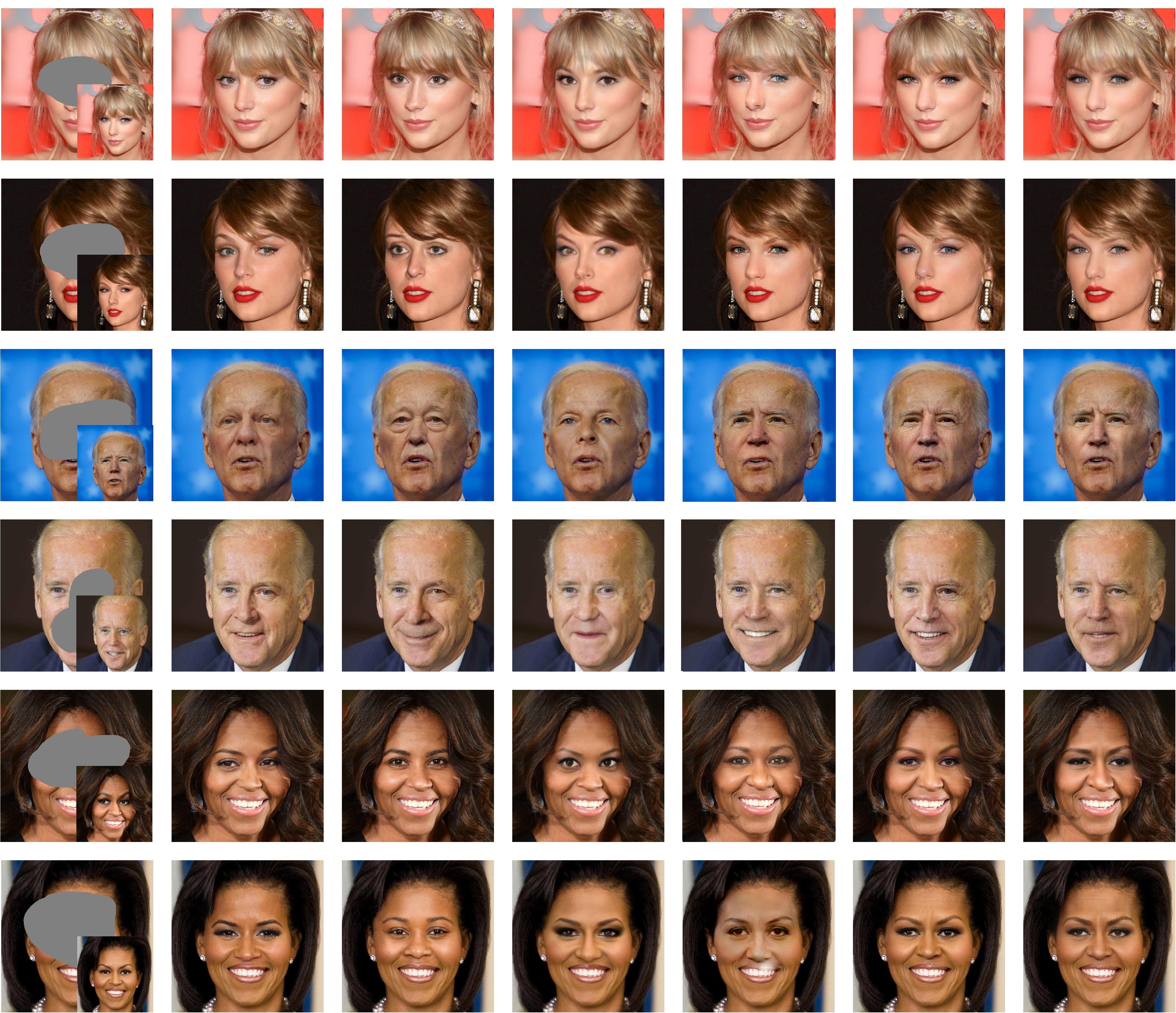}
  \caption{\label{more:res} Additional qualitative results. Please zoom in for details.}
\end{figure}

\end{document}